\documentclass[10pt,twocolumn,letterpaper]{article}

\usepackage{cvpr}              
\usepackage{graphicx}
\usepackage{multirow}
\usepackage{amssymb}
\newcommand{\tabincell}[2]{\begin{tabular}{@{}#1@{}}#2\end{tabular}}

\usepackage[dvipsnames]{xcolor}

\definecolor{cvprblue}{rgb}{0.21,0.49,0.74}
\usepackage[pagebackref,breaklinks,colorlinks,citecolor=cvprblue]{hyperref}

\def\paperID{4915} 
\def\confName{CVPR}
\def\confYear{2024}

\title{OneTracker: Unifying Visual Object Tracking with \\ Foundation Models and Efficient Tuning}

\author{Lingyi Hong\textsuperscript{\rm 1}\thanks{Equal Contribution} ~
        Shilin Yan\textsuperscript{\rm 1}\footnotemark[1] ~ 
        Renrui Zhang\textsuperscript{\rm 2} ~
        Wanyun Li\textsuperscript{\rm 1} ~
        Xinyu Zhou\textsuperscript{\rm 1} ~
        Pinxue Guo\textsuperscript{\rm 3} ~
        \\
        Kaixun Jiang\textsuperscript{\rm 3} ~ 
        Yiting Chen\textsuperscript{\rm 1} ~
        Jinglun Li\textsuperscript{\rm 3} ~ 
        Zhaoyu Chen\textsuperscript{\rm 3} ~
        Wenqiang Zhang\textsuperscript{\rm 1,4}\thanks{Corresponding Author}\footnotemark[2]  ~
        \\
        \textsuperscript{\rm 1}~Shanghai Key Lab of Intelligent Information Processing, \\School of Computer Science, Fudan University, Shanghai, China\\
        \textsuperscript{\rm 2}~The Chinese University of Hong Kong\\
        \textsuperscript{\rm 3}~Shanghai Engineering Research Center of AI \& Robotics, \\Academy for Engineering \& Technology, Fudan University, Shanghai, China\\
        \textsuperscript{\rm 4}~Engineering Research Center of AI \& Robotics, Ministry of Education, \\Academy for Engineering \& Technology, Fudan University, Shanghai, China \\
        {\tt\small \{honglyhly, tattoo.ysl\}@gmail.com, wqzhang@fudan.edu.cn}
}

\begin{document}
\maketitle
\begin{abstract}
Visual object tracking aims to localize the target object of each frame based on its initial appearance in the first frame. Depending on the input modility, tracking tasks can be divided into RGB tracking and RGB+X (e.g. RGB+N, and RGB+D) tracking. Despite the different input modalities, the core aspect of tracking is the temporal matching. Based on this common ground, we present a general framework to unify various tracking tasks, termed as \textbf{OneTracker}. OneTracker first performs a large-scale pre-training on a RGB tracker called Foundation Tracker. This pretraining phase equips the Foundation Tracker with a stable ability to estimate the location of the target object. Then we regard other modality information as prompt and build Prompt Tracker upon Foundation Tracker. Through freezing the Foundation Tracker and only adjusting some additional trainable parameters, Prompt Tracker inhibits the strong localization ability from Foundation Tracker and achieves parameter-efficient finetuning on downstream RGB+X tracking tasks.
To evaluate the effectiveness of our general framework OneTracker, which is consisted of Foundation Tracker and Prompt Tracker, we conduct extensive experiments on 6 popular tracking tasks across 11 benchmarks and our OneTracker outperforms other models and achieves state-of-the-art performance.
\end{abstract}   
\vspace{-6.4mm}
\section{Introduction}
\label{sec:intro}

\begin{figure*}
	\centering
    \includegraphics[width=\linewidth]{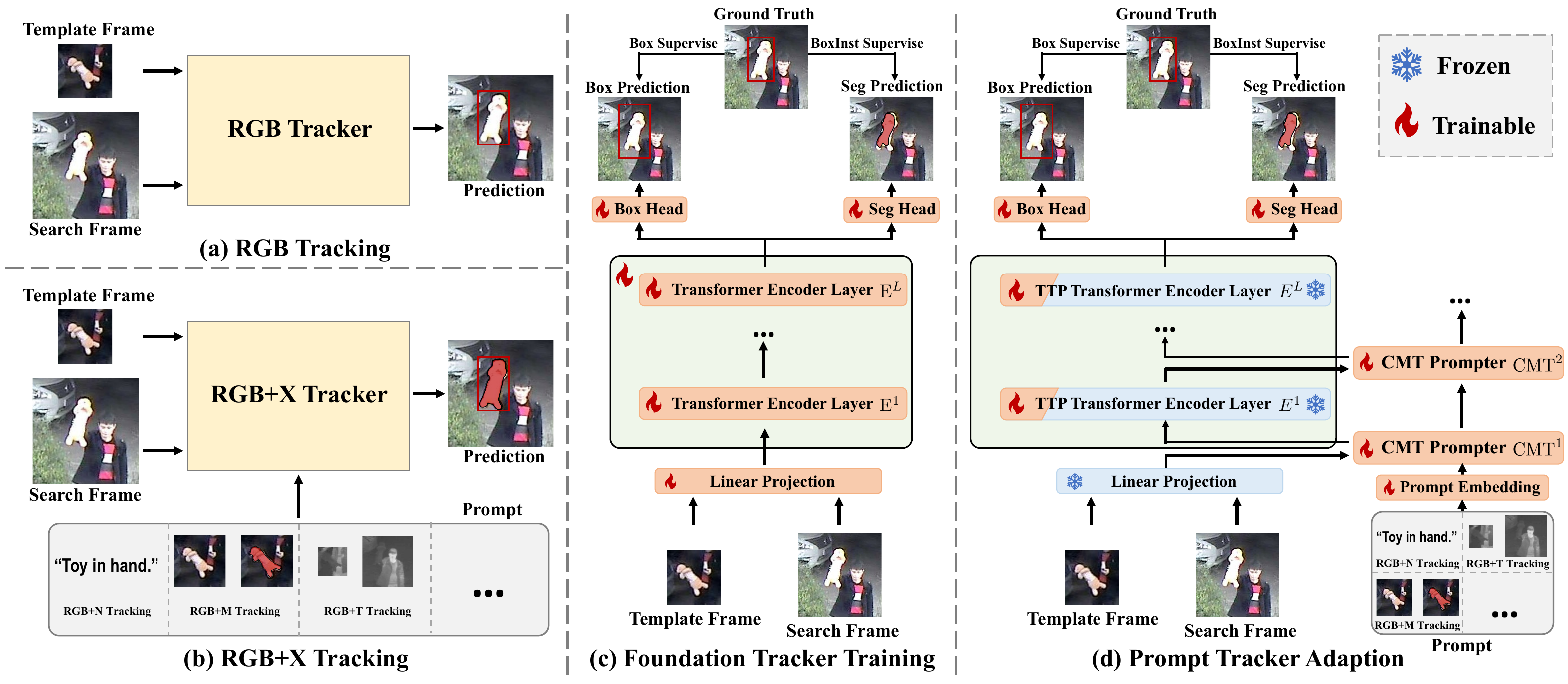}
	\caption{\textbf{(a) The definition of RGB tracking. (b) The definition of RGB+X tracking. (c) Overview of Foundation Tracker training. (d) The parameter-efficient finetuning of Prompt Tracker.}}
    \label{fig:overview}
\end{figure*}

Object tracking~\cite{bertinetto2016fully,li2019siamrpn++,wu2013online,chen2020siamese,zhou2024reading,zhou2023memory} is a foundation visual task, that involves localizing a target object in each video frame based on the initial bounding box in the first frame. It has various applications, such as self-driving~\cite{zhang2016instance,chen2015deepdriving,geiger2012we}, visual surveillance~\cite{xing2010multiple,tian2005robust}, and video compression~\cite{itti2004automatic}. In addition to the conventional RGB tracking (Figure~\ref{fig:overview} (a)), there are various downstream tracking tasks that incorporate additional information and boost performance, including RGB+N, RGB+M, and RGB+D/T/E tracking. In RGB+N tracking~\cite{li2017tracking,fan2019lasot,wang2021towards,yan2023referred}, the natural linguistic descriptions of target are additionally provided to exclude the interference of similar objects and enhance the localization. In RGB+M tracking~\cite{perazzi2016benchmark,xu2018youtube,hong2022lvos,yan2023panovos,ding2023mose}, the masks of the target in the first frame are offered instead of bounding boxes. In RGB+D/T/E tracking, the depth, thermal, and event maps are utilized as an extra to handle with the vulnerability of RGB trackers to complex scenarios and improve the robustness. 
The goal of all these downstream tasks is to localize the target with the assistance of multimodal information. Thus, we unify them as a whole, terming them as RGB+X tracking (Figure~\ref{fig:overview} (b)). Despite the diversity of tracking tasks, the core objective remains the same: localizing the target in the search frame given its initial appearance, similar to the underlying principles of human attention mechanisms~\cite{buzsaki2017space}. Cognitive scientists have discovered that the human vision system builds the correspondence or motion~\cite{bahl2013object} on the temporal dimension~\cite{kosse2019natural} to determine the object's position in the current frame~\cite{duncan1984selective}, regardless of the form of additional modalities in various tracking tasks. 

Currently, there is a prevailing trend where models are designed and trained for specific tasks using data from certain domains, offering convenience and yielding competitive results on individual tasks. However, this design philosophy presents certain challenges. (1) Independent models require customized architectures, resulting in complex training procedures and redundant parameters. (2) For certain tracking tasks, the limited availability of large-scale data severely restricts performance potential. (3) The separate design approach falls short of accurately simulating human attention mechanisms, which are crucial in tracking. Although some previous works~\cite{athar2023tarvis,yan2023universal,wang2023omnitracker,yan2022towards,ma2022unified,wang2021different} have made attempts to address these problems in a unified model, they still exhibit certain limitations. \cite{athar2023tarvis,yan2023universal} are not specifically designed for tracking tasks, resulting in sub-optimal performance on tracking benchmarks. \cite{wang2023omnitracker,yan2022towards,ma2022unified,wang2021different} only consider RGB images as input. \cite{zhu2023visual,yang2022prompting} attempt to utilize the multi-modal information, but their applicability is limited to RGB+D/T/E tracking tasks. Moreover, these models fail to capture the unified temporal attention mechanisms observed in human tracking. 

To address these challenges, we propose \textbf{OneTracker}, a general framework to unify RGB and RGB+X tracking within a consistent format. OneTracker firstly presents a Foundation Tracker for RGB tracking tasks, and then adapts it to RGB+X with parameter-efficient strategy. In detail, we pretrain a Foundation Tracker~\cite{gao2023generalized} on several RGB tracking datasets~\cite{fan2019lasot,muller2018trackingnet,huang2019got} (Figure~\ref{fig:overview} (c)). 

After the large-scale pretraining, Foundation Tracker possesses strong localization capabilities, allowing it to accurately locate the target object in the search frame based on its appearance in the template frame. Then, we proceed to finetune Foundation Tracker on specific downstream RGB+X tracking tasks, referred as Prompt Tracker (Figure~\ref{fig:overview} (d)). In contrast to leveraging an extra parallel module to fuse multimodal information, we propose Cross Modality Tracking Prompters (CMT Prompter) to introduce multimodal features in a prompt-tuning manner. CMT Prompters learn semantic understanding of multimodal information and integrate it with RGB images. Furthermore, to enhance the adaptation to downstream task, we replace the vanilla Transformer layers with Tracking Task Perception Transformer (TTP Transformer) layers. 
Because the linear layer in Transformer contains most of the knowledge of specific tasks, we only introduce few trainable parameters into each linear layer of Transformer to bridge the RGB tracking and RGB+X tracking. 
By leveraging CMT Prompter and TTP Transformer layer, Prompt Tracker inherits the strong localization ability from Foundation Tracker and achieves competitive performance on downstream RGB+X tracking tasks after a quick fine-tuning on a small number of parameters. Given the minimal number of additional parameters, Prompt Tracker maintains a similar speed to Foundation Tracker. 

\begin{figure*}
	\centering
    \includegraphics[width=\linewidth]{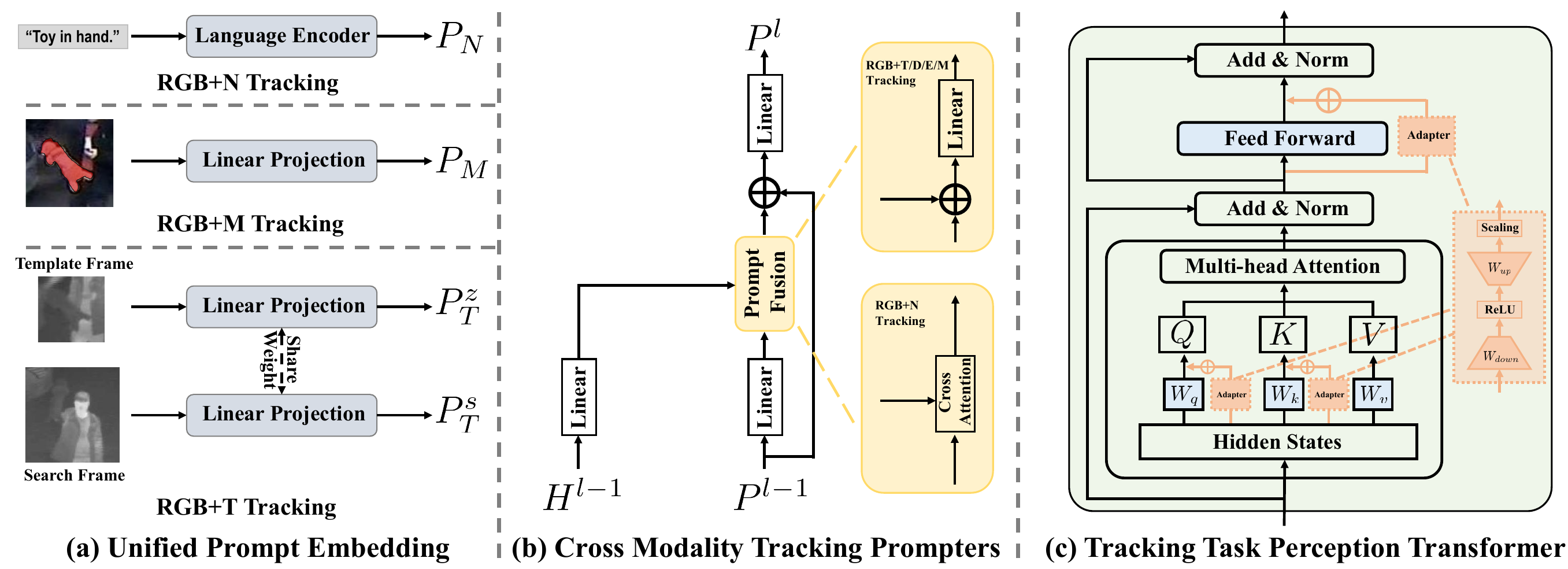}
	\caption{\textbf{(a) Unified Prompt Embedding structure. (b) Cross Modality Tracking (CMT) Prompters. (c) Tracking Task Perception (TTP) Transformer layers.}}
    \label{fig:prompt}
\end{figure*}

Overall, our contributions are summaried as follows:

\begin{itemize}
    \item We present a unified tracking architecture, termed as OneTracker, which is consisted of Foundation Tracker and Prompt Tracker, to tackle various forms of tracking tasks, i.e. both RGB tracking and RGB+N/M/D/T/E tracking.
    \item We propose a Foundation Tracker trained on several RGB tracking datasets, which owns strong ability to accurately localize target objects in search frame.
    \item To better adapt Foundation Tracker to downstream RGB+X tracking tasks efficiently, we propose CMT Prompter and TTP Transformer layer, enhancing the model's ability to incorporate additional modalities, termed as Prompt Tracker. 
    \item OneTracker achieves state-of-the-art performance on 11 benchmarks from 6 tracking tasks.
\end{itemize}

\section{Related Works}

\textbf{Large-scale Pretraining Vision Models.}
Large-scale pretraining models, or foundation models~\cite{bommasani2021opportunities}, have emerged as powerful models which are trained on broad data and can be adapted to various downstream tasks.  
These models, initially popularized in Natural Language Processing (NLP) by ~\cite{devlin2018bert,liu2019roberta,radford2019language,raffel2020exploring}, have extended their influence to multiple domains. In the realm of computer vision, \cite{zhai2022scaling} is the first to extend ViT~\cite{dosovitskiy2020image} to 2 billion parameters. \cite{bao2021beit,he2022masked, xue2023stare} learn representations from images corrupted by masking. \cite{wang2022image,su2022towards,radford2021learning, fang2023instructseq, xue2022clip} explore the vision-language training strategy to align visual and text feature in a unified space. These large-scale pretraining vision models have shown their exceptional transferability across various downstream tasks. Inspired by the success of large-scale pretraining strategies, we propose Foundation Tracker, which is trained on a combination of diverse tracking datasets and demonstrates strong temporal matching capabilities.

\textbf{Parameter-Efficient Transfer Learning.}
Parameter-efficient transfer learning (PETL) is introduced to serve as a lightweight alternative to address this limitation, which involves freezing the pretrained language model and adding a small number of extra trainable parameters to achieve quick adaptation to downstream tasks while maintaining parameter efficiency~\cite{hu2021lora,li2021prefix,liu2021p,lester2021power,houlsby2019parameter,pfeiffer2020adapterfusion}. 
PETL also demonstrates its high efficiency in computer vision fields.
VPT~\cite{jia2022visual} inserts additional parameters to the input sequence before encoder. Diverse kinds of adapter~\cite{chen2022vision,gao2021clip,zhang2021tip,chen2022adaptformer} are proposed to adjust ViT to downstream tasks.
ProTrack~\cite{yang2022prompting} and ViPT~\cite{zhu2023visual} attempt to introduce the prompting concept into trakcing area, while they just focus on RGB+D/T/E tracking. The question of how to transfer large-scale pretraining tracker to other tracking tasks, such as RGB+M and RGB+N tracking, remains unanswered. In this work, we propose Prompt Tracker based on Foundation Tracker. 
We introduce CMT Prompter and TTP Transformer layer to perform the parameter-efficient finetuing on RGB+X tracking tasks.

\textbf{Visual Tracking.} 
Visual object tracking is a fundamental task, including RGB tracking and RGB+X tracking. RGB tracking~\cite{lasot,trackingnet} utilizes raw RGB images for object tracking. 
Because only levaraging pure RGB image is prone to some complex scenarios, RGB+X tracking is proposed for robust tracking by incorporating multimodal information. RGB+T/E/D tracking~\cite{yan2021depthtrack,zhu2022rgbd1k,xiao2022attribute,zhang2022visible,zhang2022spiking,zhang2021object} take advantage of thermal or event flows or depth maps. RGB+N tracking~\cite{li2017tracking,fan2019lasot,wang2021towards} fuses language description with RGB images, and RGB+M tracking~\cite{perazzi2016benchmark,xu2018youtube,hong2022lvos,ding2023mose,guo2022adaptive,hong2023simulflow,hong2021adaptive} provides the mask of target in the first frame. Despite promising performance of task-specific~\cite{wu2013online,chen2020siamese,ostrack, mixformer}  or multi-task~\cite{athar2023tarvis,yan2023universal,wang2023omnitracker,yan2022towards,ma2022unified,wang2021different} trackers, these models can not simulate the human temporal matching mechanism well and lack the ability to handle multi-modal tracking tasks. In this work, we propose a general manner to unify RGB and RGB+X tracking tasks. 

\section{Methodology}

In this work, we propose OneTracker, consisting of Foundation Tracker and Prompt Tracker, to implement a unified framework for tracking tasks. The overall structure of Foundation Tracker and Prompt Tracker is in Figure~\ref{fig:overview} (c) and (d). We will illustrate the unification of tracking tasks (Sec.~\ref{sec:tracking_unify}), the structure of Foundation Tracker (Sec.~\ref{sec:foundation}), the structure of Prompt Tracker (Sec.~\ref{sec:prompt}), and details of training and finetuning (Sec.~\ref{sec:train}).

\subsection{Tracking Unification}
\label{sec:tracking_unify}
The core aspect of tracking  involves estimating the position of moving objects in each video frame based on its initial appearance. Depending on the different input format, tracking tasks can be divided into two main categories: RGB Tracking and RGB+X Tracking (Figure~\ref{fig:overview} (a) and (b)). 

\textbf{RGB Tracking.} RGB tracking is a extensively studied tracking task, focusing on tracking objects using RGB image information. Given a video sequence with the bounding box of target object in the first frame, the formula of RGB tracking is like,
{
\begin{equation}
\label{equ:fdtracker}
    B=\mathrm{FT}(I,B_{0};\theta),
\end{equation}
}
where $I$, $B_{0}$, $B$ denote the RGB frames of a video, the initial box prediction, the box predictions in the subsequent frames. $\mathrm{FT}$ is the Foundation Tracker with parameter $\theta$. 

\textbf{RGB+X Tracking.} We introduce an unified format that encompasses RGB+X tracking and RGB tracking. The formula for RGB+X tracking can be expressed as:
{
\begin{equation}
\label{equ:pttracker}
    B=\mathrm{PT}(I,B_{0},X;\theta^{'}),
\end{equation}
}
where $X$ is the additional information input of specific RGB+X tracking task and $\mathrm{PT}$ is the Prompt Tracker with parameter $\theta^{'}$. $X$ varies depending on the RGB+X tracking task. For RGB+N tracking, $X_{N}$ is the language description. For RGB+M tracking, $X_{M}$ is the mask of the target in the first frame and $B_{0}$ is not provided. For RGB+D/T/E tracking, $X_{D}$, $X_{T}$, and $X_{E}$ correspond to the depth, thermal, and event map of each frame. To further illustrate our framework, we can rewrite the Equation~\ref{equ:pttracker} as follows:
{
\begin{equation}
\label{equ:pttracker_u}
    \begin{aligned}
    B &=\mathrm{PT}(I,B_{0},X;\mathrm{TTP}(\theta))\\
      &=\mathrm{FT}(\mathrm{CMT}(I,B_{0},X);\mathrm{TTP}(\theta)),
    \end{aligned}
\end{equation}
}
where $\mathrm{TTP}$ denotes the replacement of the vanilla Transformer layers with our Tracking Task Perception (TTP) Transformer layers and $\mathrm{CMT}$ is the Cross Modality Tracking Prompters (CMT Prompters). Through our specific design, we succeed in unifying RGB tracking and RGB+X tracking in a general format.

\subsection{Foundation Tracker}
\label{sec:foundation}
The structure of Foundation Tracker (Figure~\ref{fig:overview} (c)) is similar to ViT~\cite{dosovitskiy2020image} with several transformer encoder layers, which are responsible for processing the input frames and capturing their spatial and temporal dependencies. To begin, the template frame $I_{z}$ and search frame $I_{s}$ are taken as input to Foundation Tracker. The RGB frames $I_{z}$ and $I_{s}$ are divided into patches and flattened into 1D tokens $H_{RGB}^{z} \in \mathbb{R}^{N_{z}\times D}$ and $H_{RGB}^{s} \in \mathbb{R}^{N_{s}\times D}$, where $N_{z}$ and $N_{s}$ denote the token number of template and search frame, and $D$ is the dimension of tokens. Then the tokens are concatenated into $H^{0}=H_{RGB}^{0}=[H_{RGB}^{z},H_{RGB}^{s}]$ and fed into $L$-layer transformer encoder layers. The forward process of the transformer encoder layers can be written as:
{
\begin{equation}
\label{equ:tr_layer}
    H^{l}=\mathrm{E}^{l}(H^{l-1}),  l = \{1,2,...,L\},
\end{equation}
}
where $\mathrm{E}^{l}$ is the $l$-th transformer encoder layers, and $H^{l-1}$ is the input to $\mathrm{E}^{l}$. The structure of transformer encoder layer is the same as the vanilla transformer layer~\cite{vaswani2017attention}.
We extract features and build the temporal matching between template and search frame. Finally, a box head is leveraged to convert the temporal correlation from transformer encoder into localization coordinates. Moreover, an extra segmentation head is leveraged to generate the mask prediction, whose structure is the same as~\cite{yang2021associating}. 
We train Foundation Tracker on several RGB tracking benchmarks, including LaSOT~\cite{fan2019lasot}, TrackingNet~\cite{muller2018trackingnet}, and GOT-10K~\cite{huang2019got}. The segmentation head is optimized in a box-supervised manner~\cite{tian2021boxinst} because of the absence of mask ground truth in tracking datasets. Similar to the foundation model in NLP, after the large-scale pretraining, our Foundation Tracker obtains the strong ability of temporal matching and transferability to downstream RGB+X tracking tasks.

\begin{table*}[htbp]
	\centering
    
	\renewcommand\arraystretch{1.15} 
	\setlength{\tabcolsep}{1.5mm}{ 
		\resizebox{\linewidth}{!}{
			\begin{tabular}{cc|cccccccccccc|c}
                \toprule[2pt]
                				
                \normalsize				
                &\multicolumn{14}{c}{\textbf{RGB Tracking}} \\
				\midrule

                \footnotesize
				&&\tabincell{c}{TransT\\~\cite{chen2021transformer}}&\tabincell{c}{STARK\\~\cite{yan2021learning}}&\tabincell{c}{MixFormer\\~\cite{cui2022mixformer}}&\tabincell{c}{OSTrack\\~\cite{ye2022joint}}&\tabincell{c}{AiATrack\\~\cite{gao2022aiatrack}}&\tabincell{c}{SimTrack\\~\cite{chen2022backbone}}&\tabincell{c}{GRM\\~\cite{gao2023generalized}}&\tabincell{c}{UniTrack\\~\cite{wang2021different}}&\tabincell{c}{UTT\\~\cite{ma2022unified}}&\tabincell{c}{Unicorn\\~\cite{yan2022towards}}&\tabincell{c}{OmniTracker\\~\cite{wang2023omnitracker}}&\tabincell{c}{UNINEXT\\~\cite{yan2023universal}}&\tabincell{c}{\textbf{One}\\\textbf{Tracker}}\\
				\midrule
                \multirow{3}{*}{\tabincell{c}{LaSOT\\~\cite{fan2019lasot}}}&AUC($\uparrow$)&64.9&66.4&69.2&69.1&69.0&69.3&69.9&35.1&64.6&65.3&69.1&69.2&\textbf{70.5} \\
	            &$\mathrm{P}_{Norm}$($\uparrow$)&73.8&76.3&78.7&78.7&79.4&78.5&79.3&-&-&73.1&77.3&77.1&\textbf{79.9} \\ &P($\uparrow$)&69.0&71.2&74.7&75.2&73.8&74.0&75.8&32.6&67.2&68.7&75.4&75.5&\textbf{76.5} \\
				\midrule


				\footnotesize
				
                \multirow{3}{*}{\tabincell{c}{TrackingNet\\~\cite{muller2018trackingnet}}}&AUC($\uparrow$)&81.4&81.3&83.1&83.1&82.7&82.3&\textbf{84.0}&-&79.7&79.0&83.4&83.2&83.7 \\
	            &$\mathrm{P}_{Norm}$($\uparrow$)&86.7&86.1&88.1&87.8&87.8&86.5&\textbf{88.7}&-&-&82.0&86.7&86.9&88.4 \\ 
                &P($\uparrow$)&80.3&78.1&81.6&82.0&80.4&-&\textbf{83.3}&-&77.0&77.4&82.3&\textbf{83.3}&82.7 \\
				\midrule[2pt]
				
                \normalsize
				&\multicolumn{14}{c}{\textbf{RGB+N Tracking}} \\
				\midrule
				\footnotesize
				&&\tabincell{c}{TNLS-III\\~\cite{li2017tracking}} &\tabincell{c}{RTTNLD\\~\cite{feng2020real}} &\tabincell{c}{SiamRPN\\~\cite{li2018high}} &\tabincell{c}{VITAL\\~\cite{song2018vital}} &\tabincell{c}{MDNet\\~\cite{nam2016learning}} &\tabincell{c}{ATOM\\~\cite{danelljan2019atom}} &\tabincell{c}{DiMP\\~\cite{bhat2019learning}}&\tabincell{c}{PrDIMP\\~\cite{danelljan2020probabilistic}}   &\tabincell{c}{SiamRPN++\\~\cite{li2019siamrpn++}}  &\tabincell{c}{TNL2K-2\\~\cite{wang2021towards}} &\tabincell{c}{SNLT\\~\cite{feng2021siamese}} &\tabincell{c}{JointNLT\\~\cite{zhou2023joint}}&\tabincell{c}{\textbf{One}\\\textbf{Tracker}}\\
				
				\midrule
				\multirow{2}{*}{\tabincell{c}{OTB99\\~\cite{li2017tracking}}}&AUC($\uparrow$) &55.0 &61.0 &61.2 &65.2 &64.6 &67.6 &67.3 &68.3  &65.8  &68.0 &66.6 &65.3 & \textbf{69.7} \\
				&P($\uparrow$)&72.0 &79.0 &75.8  &84.2 & 82.8 &82.4 &81.9 &83.0 & 79.7  &88.0 &80.4 &85.6 &\textbf{91.5} \\
				\midrule
				
				\footnotesize
				
				\multirow{2}{*}{\tabincell{c}{TNL2K\\~\cite{wang2021towards}}}&AUC($\uparrow$) &- &25.0 &- &- &- &- &- &-  &- &42.0 &27.6 &56.9 &\textbf{58.0} \\
				&P($\uparrow$)&-&27.0 &-  &- & - &- &- &- & - &42.0 &41.9 &58.1 & \textbf{59.1} \\
				\midrule[2pt]

                \normalsize	
                &\multicolumn{14}{c}{\textbf{RGB+D Tracking}} \\
                \midrule
				\footnotesize
				&&\tabincell{c}{ATOM\\~\cite{atom}}&\tabincell{c}{LTDSEd\\~\cite{vot19}}&\tabincell{c}{DRefine\\~\cite{vot21}}&\tabincell{c}{keep\_track\\~\cite{vot22}}&\tabincell{c}{LTMU\_B\\~\cite{vot20}}&\tabincell{c}{DiMP\\~\cite{dimp}}&\tabincell{c}{DDiMP\\~\cite{vot20}}&\tabincell{c}{DeT\\~\cite{depthtrack}}&\tabincell{c}{OSTrack\\~\cite{ostrack}}&\tabincell{c}{SPT\\~\cite{rgbd1k}}&\tabincell{c}{ProTrack\\~\cite{protrack}}&\tabincell{c}{ViPT\\~\cite{zhu2023visual}}&\tabincell{c}{\textbf{One}\\\textbf{Tracker}}\\
				\midrule
				\multirow{3}{*}{\tabincell{c}{DepthTrack\\~\cite{yan2021depthtrack}}}&F-score($\uparrow$)&-&40.5&-&-&46.0&-&48.5&53.2&52.9&53.8&57.8&59.4 & \textbf{60.9} \\
				&R($\uparrow$)&-&38.2&-&-&41.7&-&46.9&50.6&52.2&54.9&57.3&59.6 & \textbf{60.4} \\
				&P($\uparrow$)&-&43.0&-&-&51.2&-&50.3&56.0&53.6&52.7&58.3&59.2 & \textbf{60.7}\\
				\midrule

				\footnotesize
				\multirow{3}{*}{\tabincell{c}{VOT\\RGBD2022\\~\cite{vot22}}}&EAO($\uparrow$)&50.5&-&59.2&60.6&-&54.3&-&65.7&67.6&65.1&65.1&72.1&\textbf{72.7} \\
				&Accuracy($\uparrow$)&69.8&-&77.5&75.3&-&70.3&-&76.0&80.3&79.8&80.1&81.5&\textbf{81.9} \\
				&Robustness($\uparrow$)&68.8&-&76.0&79.7&-&73.1&-&84.5&83.3&85.1&80.2&87.1&\textbf{87.2} \\
				\midrule[2pt]
				
                \normalsize	
				&\multicolumn{14}{c}{\textbf{RGB+T Tracking}} \\
				\midrule
				\footnotesize
				&&\tabincell{c}{SGT++\\~\cite{sgt}} &\tabincell{c}{DAPNet\\~\cite{zhu2019dense}} &\tabincell{c}{HMFT\\~\cite{zhang2022visible}} &\tabincell{c}{FANet\\~\cite{zhu2020quality}}&\tabincell{c}{mfDiMP\\~\cite{zhang2019multi}}  &\tabincell{c}{STARKS50\\~\cite{yan2021learning}} &\tabincell{c}{CAT\\~\cite{cat}} &\tabincell{c}{APFNet\\~\cite{apfnet}} &\tabincell{c}{OSTrack\\~\cite{ye2022joint}} &\tabincell{c}{TransT\\~\cite{chen2021transformer}} &\tabincell{c}{ProTrack\\~\cite{protrack}} &\tabincell{c}{ViPT\\~\cite{zhu2023visual}}&\tabincell{c}{\textbf{One}\\\textbf{Tracker}}\\
				
				\midrule
				\multirow{2}{*}{\tabincell{c}{LasHeR\\~\cite{lasher}}}&PR($\uparrow$) &36.5 &43.1 &43.6 &44.1 &44.7 &44.9 &45.0 &50.0 &51.5 &52.4 &53.8 &65.1 & \textbf{67.2} \\
				&SR($\uparrow$) &25.1 &30.9 &31.3 &31.4 &31.4 &34.3 &36.1 &36.2 &39.4 &41.2 &42.0 &52.5 & \textbf{53.8} \\
				\midrule
				
				\footnotesize
				
				\multirow{2}{*}{\tabincell{c}{RGBT234\\~\cite{rgbt234}}}&MPR($\uparrow$) &64.6 &72.0 &79.6 &72.9 &78.7 &79.0 &80.4 &79.0 &82.3 &82.7 &79.5 &83.5 & \textbf{85.7} \\
				&MSR($\uparrow$) &42.8 &47.2 &54.4 &54.9 &55.3 &55.4 &56.1 &57.3 &57.5 &57.9 &59.9 &61.7 &\textbf{64.2} \\
				\midrule[2pt]
				
                \normalsize	
				&\multicolumn{14}{c}{\textbf{RGB+E Tracking}} \\
				\midrule
				\footnotesize
				&&\tabincell{c}{MetaTracker\\~\cite{park2018meta}} &\tabincell{c}{ATOM\\~\cite{atom}} &\tabincell{c}{STARKS50\\~\cite{yan2021learning}} &\tabincell{c}{ProTrack\\~\cite{protrack}}&\tabincell{c}{PrDIMP50\\~\cite{danelljan2020probabilistic}}  &\tabincell{c}{VITAL\\~\cite{song2018vital}} &\tabincell{c}{TransT\\~\cite{chen2021transformer}} &\tabincell{c}{LTMU\\~\cite{dai2020high}} &\tabincell{c}{SiamRCNN\\~\cite{voigtlaender2020siam}} &\tabincell{c}{MDNet\\~\cite{nam2016learning}} &\tabincell{c}{OSTrack\\~\cite{ye2022joint}} &\tabincell{c}{ViPT\\~\cite{zhu2023visual}}&\tabincell{c}{\textbf{One}\\\textbf{Tracker}}\\
				
				\midrule
				\multirow{2}{*}{\tabincell{c}{VisEvent\\~\cite{visevent}}}&MPR($\uparrow$) &49.1 &60.8 &61.2 &63.2 &64.4 &64.9 &65.0 &65.5 &65.9 &66.1 &69.5 &75.8 & \textbf{76.7}\\
				&MSR($\uparrow$) &29.8 &41.2 &44.6 &47.1 &45.3 & -  &47.4 &45.9 &49.9 & - &53.4 &59.2 & \textbf{60.8}\\
				\midrule[2pt]
				
				&\multicolumn{14}{c}{\textbf{RGB+M Tracking}} \\
				\midrule
				\footnotesize
				&&\tabincell{c}{STM\\~\cite{oh2019video}}&\tabincell{c}{CFBI\\~\cite{yang2020collaborative}}&\tabincell{c}{AOT\\~\cite{yang2021associating}}&\tabincell{c}{STCN\\~\cite{cheng2021rethinking}}&\tabincell{c}{XMem\\~\cite{cheng2022xmem}}&\tabincell{c}{SiamMask\\~\cite{siammask}}&\tabincell{c}{Siam R-CNN\\~\cite{voigtlaender2020siam}}&\tabincell{c}{UniTrack\\~\cite{wang2021different}}&\tabincell{c}{Unicorn\\~\cite{yan2022towards}}&\tabincell{c}{TarVIS\\~\cite{athar2023tarvis}}&\tabincell{c}{OmniTracker\\~\cite{wang2023omnitracker}}&\tabincell{c}{UNINEXT\\~\cite{yan2023universal}}&\tabincell{c}{\textbf{One}\\\textbf{Tracker}}\\
				
				\midrule
				\multirow{3}{*}{\tabincell{c}{DAVIS16\\~\cite{perazzi2016benchmark}}}&$\mathcal{J \& F}$($\uparrow$) &89.3 &89.4&91.1& 91.6 & 92.0 &69.8 & - & - & 87.4 &- &88.5 &- & \textbf{88.9} \\
				&$\mathcal{J}$($\uparrow$) &88.7 &88.3 &90.1 & 90.8  & 90.7 & 71.7  & - & - &86.5 & - &87.3 &- & \textbf{88.1} \\
				&$\mathcal{F}$($\uparrow$) &89.9 & 90.5 &92.1 & 92.5 & 93.2 & 67.8  & - & -  &88.2 & - &89.7 &- & \textbf{89.7}\\
				\midrule
				
				\footnotesize
				
				\multirow{3}{*}{\tabincell{c}{DAVIS17\\~\cite{pont20172017}}}&$\mathcal{J \& F}$($\uparrow$) &81.8 &81.9&84.9&85.4 &86.2 &56.4 &70.6 &-&69.2 &82.0 &71.0 &81.8 & \textbf{82.5}\\
				&$\mathcal{J}$($\uparrow$) &79.2 &79.1 &82.3 &82.2 &82.9 & 54.3  &66.1 &58.4 &65.2 & 78.7 &66.8 &77.7 & \textbf{79.4}\\
				&$\mathcal{F}$($\uparrow$) &84.3 &84.6 &87.5 &88.6&89.5 & 58.5  &75.0 & -  &73.2 & 87.0 &75.2 &85.8 & \textbf{85.6} \\
				\bottomrule[2pt]

			\end{tabular} 
	}}
        \caption{\textbf{Overall performance on RGB tracking and RGB+X tracking.}}
	\label{tab:total_performance}
\end{table*}

\subsection{Prompt Tracker}
\label{sec:prompt}
Different from previous RGB+X works, which add an additional module to fuse mutlimodal features, we regard the multimodal information as a kind of prompt and provide Foundation Tracker with complementarity in a prompt-tuning manner, termed as Prompt Tracker (Figure~\ref{fig:overview} (d)). To enable efficient adaptation to downstream tasks, we propose the Cross Modality Tracking Prompters (CMT Prompters) and the Tracking Task Perception Transformer (TTP Transformer) layers.

\textbf{Unified Prompt Embedding.}  
With the general definition of RGB+X tracking, the Prompt Tracker leverages a unified prompt embedding module (Figure~\ref{fig:prompt} (a)) to transform different modality downstream information into tokens $P^{0}=P_{X}$. The choice of prompt embedding strategy depends on the specific downstream task's modality. To deal with language description in RGB+N tracking, we adopt BERT~\cite{devlin2018bert} as a language encoder to extract the linguistic feature $P_{N} \in \mathbb{R}^{L\times D}$ with a sequence length of $L$. For RGB+M tracking, a patch embed layer is utilized to project the mask of the target object into patches and flatten them into 1D tokens $P_{M} \in \mathbb{R}^{N_{z}\times D}$. The size of $P_{M}$ is the same as $H_{RGB}^{z}$. For RGB+T tracking, the corresponding multimodal maps of the template and search frame are fed into a patch embed layer and then flattened into 1D tokens $P_{T} \in \mathbb{R}^{N_{z}\times D}$ and $P_{T} \in \mathbb{R}^{N_{s}\times D}$. The prompt embedding of RGB+D and RGB+E tracking follows the same procedure as RGB+T tracking. Through the unified prompt embedding module, we effectively map the multimodal information into a cohesive token representation.

\textbf{Cross Modality Tracking Prompters.}
After unified prompt embedding, we propose the Cross Modality Tracking Prompters (CMT Prompters) to fuse the extra information. Although a few works have attempted to insert some trainable parameters into pretrained models to bridge the upstream and downstream tasks, how to integrate the cross-modal information for tracking tasks is more challenging. CMT Prompters are designed to extract the semantic representations of multimodal information and provide Foundation Tracker with complementarity. As depicted in Figure~\ref{fig:prompt} (b), CMT Prompters consist of multiple linear layers and a prompt fusion module, which can be written as:
{
\begin{equation}
\label{equ:cmt}
    P^{l+1}=\mathrm{CMT}^{l}(H^{l},P^{l}), l = \{0,1,...,L-1\},
\end{equation}
}
where $\mathrm{CMT}^{l-1}$ denotes the $l$-th CMT Prompter, and $P^{l}$ is the output of $\mathrm{CMT}^{l}$. The prompt $P^{l}$ is added to the original matching results $H^{l}$ in a form of residuals:
{
\begin{equation}
\label{equ:cmt_p}
    H^{l}=H^{l}+P^{l+1}, l = \{0,1,...,L-1\}.
\end{equation}
}

CMT Prompters take the matching results $H^{l}$ from $l$-th transformer encoder layer and the prompt $P^{l}$ as input. Firstly, the $H^{l}$ and $P^{l}$ are mapped to lower-dimensional latent space using a linear layer, respectively. Subsequently, a prompt fusion module is employed to integrate the modalities. For RGB+N tracking, cross-attention is utilized to merge the linguistic feature and temporal correlation. For other RGB+X tracking, $P^{l}$ and $H^{l}$ are added and merged by a linear layer. Finally, another linear layer projects the fused feature to the original dimension. The structure and format of the CMT Prompter remain consistent across different RGB+X tracking tasks. By leveraging CMT Prompter, we achieve the integration between RGB images and multimodal information with high efficiency through prompt-tuning techniques.

\begin{table*}[htbp]
  \centering
  \renewcommand\arraystretch{1.0} 
        \tiny
        \setlength{\tabcolsep}{1mm}{\resizebox{\linewidth}{!}{\begin{tabular}{c| c | ccc|ccc|cc|cc|cc|ccc}
	\toprule
	\multirow{2}{*}{Method} & \multirow{2}{*}{\# Params} & \multicolumn{3}{c|}{LaSOT}   & \multicolumn{3}{c|}{DepthTrack}   & \multicolumn{2}{c|}{LasHeR}  &   \multicolumn{2}{c|}{VisEvent}     & \multicolumn{2}{c|}{OTB}      & \multicolumn{3}{c}{DAVIS17}   \\
    \cmidrule(l){3-17}

	& & AUC    & $\mathrm{P}_{norm}$ & P & F    & R & P & PR     & SR & PR & SR & AUC & P & $\mathcal{J}\& \mathcal{F}$    & $\mathcal{J}$ & $\mathcal{F}$ \\
	\midrule
       
	\tabincell{c}{Foundation Tracker}& - & 70.5  &   79.9  & 76.5  &  55.9  &  55.6  &  55.7  &  53.3    &  42.1  &    70.1      & 53.6   &  67.3   &  88.9   &   42.7    &  37.4  &  48.1 \\
	Full Finetune    &  99.83M    & - &   -     & -  & 57.2     &  56.9  & 57.1   &   65.4   &  52.5   &  75.6  &  59.8  &  68.5   & 89.6    &   77.8   &  75.4  & 80.2  \\
	\tabincell{c}{\textbf{Prompt Tracker}}& 2.8M &  -   & -   & - & \textbf{60.9}    & \textbf{60.4} & \textbf{60.7} & \textbf{67.2}  & \textbf{53.8} & \textbf{76.7}  & \textbf{60.8} & \textbf{69.7} & \textbf{91.5} & \textbf{82.5}  & \textbf{79.4} & \textbf{85.6} \\
	\tabincell{c}{w/o CMT Prompters}& 2.55M & -   & -   & - & 56.5  & 55.4 & 56.7 & 60.7  & 47.1 & 74.0  & 54.5 & 68.7 & 89.9 & 80.4 & 78.6 & 82.2 \\
	\tabincell{c}{w/o TTP TransFormer}& 0.25M & -   & -   & - & 59.2    & 58.8 & 59.1 & 65.6  & 52.3 & 75.3 & 59.0 & 69.3 & 90.8 & 81.7  & 79.3 & 84.0 \\
	
	\bottomrule
\end{tabular} }}
        \caption{\textbf{Ablation study on the Prompt Tracker.} \# Params denotes the number of trainable parameters.}
    \label{tab:ab_foundation}
\end{table*}

\textbf{Tracking Task Perception Transformer.}
Although CMT Prompter effectively complements auxiliary modalities as prompts, the Prompt Tracker lacks specialization for certain downstream tasks. For example, Foundation Tracker excels at localizing targets based on RGB images, while the lack of perception of linguistic features may result in suboptimal performance in RGB+N tracking tasks. Given the Foundation Tracker parametrazed by $\theta$, $\theta$ may not be the optimal weights for downstream tasks. Suppose the best weights on downstream tasks are $\theta^{'}$, the purpose of full finetuning is to learn difference $\Delta\theta$ between $\theta$ and $\theta^{'}$ and update Foundation Tracker to $\theta + \Delta\theta$. The drawbacks of full finetuning are that we must learn a different set of parameters $\Delta\theta$, whose dimension $|\Delta\theta|$ is equal to $|\theta|$ for each different downstream task, and lack of large-scale data in specific downstream tasks may lead to suboptimal finetuning performance and result in catastrophic forgetting.

Thus, to bridge the gap between RGB tracking and downstream RGB+X tracking tasks, we propose Tracking Task Perception Transformer (TTP Transformer) by adding some adapters to Foundation Tracker. Where to insert trainable parameters is a crucial question. The linear layer in transformer encoder contains the knowledge of specific tasks, especially the linear layer in Feed Forward Network (FFN)~\cite{he2021towards}. As shown in Figure~\ref{fig:prompt} (c), we insert trainable adapters with a small number of parameters to the linear projection operation in vanilla transformer encoder layers, i.e. the query/key/value projection matrixes and the output layers in FFN to enable the efficient adaption. The structure of the adapter follows a similar approach with ~\cite{he2021towards}. For a pretrained linear layer with weight matrix $W\in \mathbb{R}^{d\times k}$, the formula can be written as:
{
\begin{equation}
\label{equ:linear}
    h=Wx,
\end{equation}
}
where $h \in \mathbb{R}^{d\times t}$ and $x \in \mathbb{R}^{k\times t}$ denote the output and input. $k$ and $d$ are the dimension of $h$ and $x$. $t$ is the token number of $x$. With an adapter, the process becomes:
{
\begin{equation}
\label{equ:lora_linear}
    h=Wx+\Delta Wx=Wx+s\cdot W_{up}\mathrm{ReLU}(W_{down}x),
\end{equation}
}
where $W_{down} \in \mathbb{R}^{d\times r}$ and $W_{up} \in \mathbb{R}^{r\times d}$ is two mapping matrixes, $\mathrm{ReLU}$ is relu operation, $s$ is the constant scaling factor, and rank $r \ll min(d,k)$. During finetuning, we freeze the $W$ and only update $W_{down}$ and $W_{up}$. Through optimizing $W_{down}$ and $W_{up}$, we achieve the highly efficient learning of $\Delta\theta$. The TTP Transformer layers bridge the RGB tracking and RGB+X tracking while maintaining the temporal matching knowledge in Foundation Tracker.

\subsection{Training and Inference} 
\label{sec:train}
\textbf{Training.}
The whole training process of OneTracker consists of two stages: Foundation Tracker pretraining and Prompt Tracker finetuning. In the first pretraining stage, we pretrain our Foundation Tracker on a combination of several large-scale RGB tracking datasets, including LaSOT~\cite{fan2019lasot}, TrackingNet~\cite{muller2018trackingnet}, and GOT-10K~\cite{huang2019got}, which is the same as previous trackers~\cite{ye2022joint,gao2023generalized,mixformer}. Following ~\cite{ye2022joint,gao2023generalized,mixformer}, we adopt the weighted focal loss~\cite{law2018cornernet} for classification, $l_{1}$ loss and generalized IoU loss~\cite{rezatofighi2019generalized} for bounding box regression. Because there is no mask annotations in tracking datasets, we leverage BoxInst~\cite{tian2021boxinst} to supervise the segmentation head of Foundation Tracker. The total loss function can be formulated as:
{
\begin{equation}
	\label{equ:stg1_loss}
	L_{stage1}=L_{cls}+\lambda_{iou}L_{iou}+\lambda_{L_{1}}L_{1}+\lambda_{L_{mask}}L_{mask}^{boxinst}
\end{equation}
}

In the second stage, we finetune our Foundation Tracker on RGB+X downstream tracking datasets. We freeze the parameters of Foundation Tracker and only train the CMT Prompters and adapters in TTP Transformer layers. For RGB+ N/D/T/E tracking, the loss function $L_{stage2}$ is equal to $L_{stage1}$. For RGB+M tracking, due to the available mask annotations, we drop the BoxInst auxiliary loss and utilize the mask annotations to optimize the segmentation head.

\textbf{Inference.}
Because of the slight difference in the input format of several tracking tasks, we adopt different inference manners. For RGB tracking and RGB+N tracking, Hanning window penalty is utilized to leverage the positional prior following previous works~\cite{gao2023generalized,ye2022joint,mixformer}. 
For RGB+D/T/E tracking, the multimodal map is also cropped by using Hanning window penalty. For RGB+M tracking, the first frame with the mask annotation and the previous frame with the predicted mask are fed into the Prompt Tracker to perform online target matching without cropping. Due to the specific design of OneTracker, we can apply them to several tracking tasks without any modification to the structure of models.

\section{Experiments}

\begin{table*}[htbp]
  \centering
  \renewcommand\arraystretch{1.0} 
        \tiny
        \resizebox{\linewidth}{!}{\begin{tabular}{c|c|  ccc|cc|cc|cc|ccc}
	\toprule
	\multirow{2}{*}{Number}& \multirow{2}{*}{\# Params} & \multicolumn{3}{c|}{DepthTrack}   & \multicolumn{2}{c|}{LasHeR}  &   \multicolumn{2}{c|}{VisEvent}     & \multicolumn{2}{c|}{OTB}      & \multicolumn{3}{c}{DAVIS17}   \\
	  \cmidrule(l){3-14}
    & & F   &  R & P & PR  & SR & PR   & SR & AUC & P & $\mathcal{J}\& \mathcal{F}$    & $\mathcal{J}$ & $\mathcal{F}$ \\
	\midrule
	0  &  -   &     56.5 & 58.8  & 59.1 & 60.7  & 47.1 & 74.0    & 54.5 &  68.7 &   89.9  &  80.4  & 78.6 & 82.2   \\
	1  & 0.02M & 57.6     & 57.2 & 57.3 & 61.5  & 48.7 & 75.7    & 59.2 &  68.9    &  90.0     &    \textbf{82.5}   &   \textbf{79.4}    &   \textbf{85.6}   \\
	2  & 0.04M & 58.4      & 58.1 & 58.2 & 63.2  &  50.1  &  76.0  &  59.5 & 69.2   &  90.4    &   75.4     &   60.7   & 77.5     \\
	4  & 0.08M & 59.1      & 58.9 & 59.3 & 64.1  & 51.0 & 76.1    & 59.7 &   69.3   &   90.4    &   67.1    &  73.3   &  73.5   \\
	6  & 0.12M & 59.5     & 59.3 & 59.4 & 65.7  & 52.5 & 76.4    & 60.3 &   69.5   &   91.0    &   58.7    &  52.5    &  64.9    \\
	12 & 0.25M & \textbf{60.9}      & \textbf{60.4} & \textbf{60.7} & \textbf{67.2}  & \textbf{53.8} & \textbf{76.7}    & \textbf{60.8} & \textbf{69.7} & \textbf{91.5} & 48.3  & 44.5 & 52.1 \\
	
	\bottomrule
\end{tabular} }
         \caption{\textbf{Ablation study on the number of CMT Prompters.} \# Params denotes the number of trainable parameters. In this experiment, we just count the number of parameters in CMT Prompters.}
    \label{tab:ab_cmt_layer}
\end{table*}

\begin{table*}[htbp]
  \centering
  \renewcommand\arraystretch{1.0} 
       \tiny
        \resizebox{\linewidth}{!}{\begin{tabular}{ccc|ccc|cc|cc|cc|ccc}
	\toprule
	\multirow{2}{*}{Task} & \multirow{2}{*}{CMT} & \multirow{2}{*}{TTP} & \multicolumn{3}{c|}{DepthTrack}   & \multicolumn{2}{c|}{LasHeR}  &   \multicolumn{2}{c|}{VisEvent}     & \multicolumn{2}{c|}{OTB}      & \multicolumn{3}{c}{DAVIS17}   \\

	 \cmidrule(l){4-15}
    & & & F  & R & P & PR  & SR & PR  & SR & AUC & P & $\mathcal{J}\& \mathcal{F}$    & $\mathcal{J}$ & $\mathcal{F}$ \\
	\midrule
	& &   &     53.9  & 53.2 & 53.4    & 59.7  & 48.5 & 73.5    & 56.8 &  68.4  &  89.6    &  58.7     &  51.1    &  66.3    \\
	&  \checkmark &   &   55.4  & 54.4  & 54.9  & 62.8 & 50.2 & 74.3  & 57.7 &  68.9    &   90.3    &    65.3   &  60.2    &  70.4    \\
	& &  \checkmark  & 57.9  & 57.0 & 57.2 &  64.8 & 52.0 & 74.8  &  58.4  &  69.2    &   90.5    &  70.2   &  66.8    &  73.6    \\
	&  \checkmark &  \checkmark  & 58.6  & 58.1 & 57.9 & 66.8  & 53.2 & 74.9  & 58.8 &  69.6    &   91.1    &   76.4    &  73.5    &   79.3   \\
	\checkmark & \checkmark &  \checkmark  & \textbf{60.9}  & \textbf{60.4} & \textbf{60.7} & \textbf{67.2}  & \textbf{53.8} & \textbf{76.7}  & \textbf{60.8} &  \textbf{69.7}  &  \textbf{91.5}   &  \textbf{82.5}  & \textbf{79.4}   &  \textbf{85.6}  \\

        \bottomrule
\end{tabular} }
        \caption{\textbf{Ablation study on the training strategy.}}
        \label{tab:ab_train}
\end{table*}

\begin{figure*}
   
	\centering
    \includegraphics[width=\linewidth]{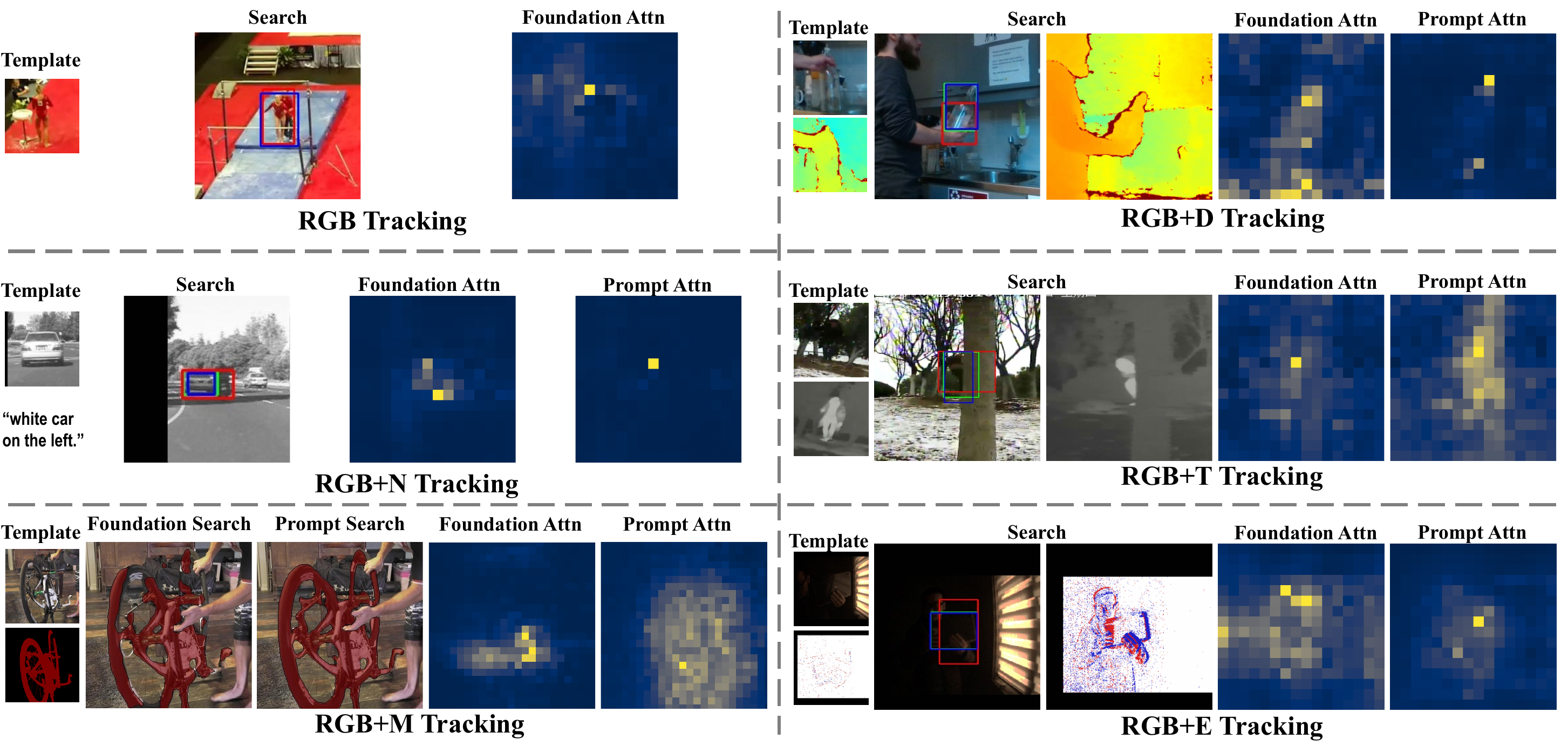}
	\caption{\textbf{Visualization results.} The \textcolor[RGB]{0,0,255}{\textbf{blue}}, \textcolor[RGB]{255,0,0}{\textbf{red}}, and \textcolor[RGB]{0,255,0}{\textbf{green}} bounding boxes denote ground truth, Foundation Tracker, and Prompt Tracker. Foundation Attn and Prompt Attn denotes the attention map of Foundation Tracker and Prompt Tracker. }
    \label{fig:vis}
\end{figure*}

\subsection{Implementation Details}
OneTracker is built on the encoder of ViT-B~\cite{vit}, which includes 12 sequential transformer layers.  The box head and segmentation head follow the structure in~\cite{ostrack,mixformer,gao2023generalized} and ~\cite{yang2021associating}, respectively. For RGB+N tracking tasks, we adopt BERT~\cite{devlin2018bert} as text encoder. During the first pretraining stage, Foundation Tracker is optimized with AdamW optimizer~\cite{loshchilov2017decoupled} for 300 epochs. 
The initial learning rate is $4 \times 10^{-5}$ for the ViT backbone and $4 \times 10^{-4}$ for the heads. It decays by a factor of 10 after 240 epochs. For the finetuning of Prompt Tracker, we freeze the parameters of Foundation Tracker and only adapt the CMT Prompter and TTP Transformer layers. We finetune Prompt Tracker for 60 epochs on the corresponding training data of each downstream tasks with an initial learning rate of $4 \times 10^{-5}$ by using AdamW optimizer, and learning rate decreased by 10 after 48 epochs. We set $\lambda_{iou}$ as 2, $\lambda_{L_{1}}$ as 5, $\lambda_{L_{mask}}$ as 1, and $r$ as 16. More details are in supplementary materials.

\subsection{Benchmark Results on 6 Tracking Tasks}
We evaluate our OneTracker on 6 tracking tasks, including RGB tracking and RGB+X Tracking. We compare the results with task-specific counterparts in Table~\ref{tab:total_performance}.

\textbf{RGB Tracking.} To show the strong temporal matching ability of our OneTracker, we compare our Foundation Tracker on widely-used RGB tracking benchmarks: LaSOT~\cite{fan2019lasot} and TrackingNet~\cite{muller2018trackingnet}. Area under the success curve (AUC), normalized precision ($P_{Norm}$), and precision (P) are adopted as metrics. Our model achieves 70.5 AUC and 69.7 AUC on LaSOT and TrackingNet, respectively, outperforming all other trackers. UniTrack, UTT, Unicorn, and OmniTracker are designed for multiple RGB tracking tasks, and UNINEXT is good at instance perception. Our Foundation Tracker surpasses these models at least 1.3 AUC on LaSOT. GRM is one of the strongest tracking-specific model, and Foundation Tracker also outperforms it 0.6 AUC on LaSOT. Moreover, our Foundation Tracker can generate the mask of the target object due to an extra segmentation head, which is not possible with other trackers.

\textbf{RGB+N Tracking.} Following previous works~\cite{zhou2023joint}, we conduct experiments on OTB99 and TNL2K. Our Prompt Tracker superpasses all existing RGB+N trackers at least 1.7 AUC and 2.5 precision on OTB99, although Prompt Tracker is not specifically designed for RGB+N tracking.

\textbf{RGB+D/T/E Tracking.} Following ~\cite{zhu2023visual}, we evalute our Prompt Tracker on DepthTrack~\cite{depthtrack} and VOT-RGBD2022~\cite{vot22} for RGB+D tracking, LasHeR~\cite{lasher} and RGBT234~\cite{rgbt234} for RGB+T tracking, and VisEvent~\cite{visevent} for RGB+E tracking.
Our Prompt Trakcer greatly exceeds all other trackers in terms of performance. Despite the fact that ViPT~\cite{zhu2023visual} also adopts a similar prompt-tuning structure, our superior results demonstrate the effectiveness of CMT Prompters and TTP Transformer layers.

\textbf{RGB+M Tracking.} We choose DAVIS16~\cite{perazzi2016benchmark} and DAVIS17~\cite{pont20172017} for RGB+M tracking. DAVIS16 is a single-object benchmark with 20 evaluation splits and DAVIS17 is the multi-object expansion of DAVIS16. Region similarity $\mathcal{J}$, contour accuracy $\mathcal{F}$, and their averaged score $\mathcal{J} \& \mathcal{F}$ are adopted as metrics. Our Prompt Tracker achieves the best performance on its multi-task counterparts~\cite{yan2023universal,athar2023tarvis} and other unified tracking models~\cite{wang2023omnitracker,yan2022towards,wang2021different,siammask,voigtlaender2020siam} by a large margin on both DAVIS16 and DAVIS17, despite its less training data and training cost. There still exists a small gap between Promp Tracker and specific models with memory mechanisms~\cite{cheng2021rethinking,cheng2022xmem,yang2021associating}, but our Prompt Tracker only relies on the first and previous frame, enabling it to handle videos with any length.

\subsection{Ablation Study.}

\textbf{Foundation Tracker and Prompt Tracker.}  To verify the strong temporal matching ability of Foundation Tracker and the effectiveness of parameter-efficient finetuning, we evaluate the performance on RGB and RGB+X tracking in Table~\ref{tab:ab_foundation}. In the first row, we solely feed the RGB image into Foundation Tracker, and the results demonstrate its strong ability to track based on visual information alone. Then, we conduct full finetuning of the Foundation Tracker on downstream RGB+X tracking datasets, as well as parameter-efficient finetuning using our proposed CMT Prompters and TTP Transformer layers. The integration of multi-modal information boosts the localization accuracy, while our Prompt Tracker achieves better performance while only adjusting 2.8M parameters, which highlights the effectiveness of our Foundation Tracker and Prompt Tracker. Furthermore, we perform an ablation study on the CMT Prompters and TTP Transformer layers. The results in the fourth and fifth rows illustrate the impact of these components in enhancing the tracking performance.

\textbf{CMT Prompter Layers.} We explore the impact on the performance of different numbers of CMT Prompter layers in Table~\ref{tab:ab_cmt_layer}. We insert CMT Prompters at different positions, each 1, 2, 3, 6, 12 transformer blocks. A value of 0 for CMT Prompter layer denotes that we add the embedding of multimodal information to RGB image embeddings directly. As the number of CMT Prompter layers increases, the performance of Prompt Tracker improves, suggesting the effectiveness of our CMT Prompters. However, interestingly, the performance on RGB+M tracking shows the opposite trend, with a significant drop in performance as the number of CMT Prompter layers increases. This observation demonstrates that the mask embedding is effective only in capturing superficial features and does not provide substantial benefits when incorporated deeply into the model.

\textbf{Training Strategy.} We analyze the training strategy of our Prompt Tracker on RGB+X tracking tasks in Table~\ref{tab:ab_train}. We investigate different approaches to jointly training the Prompt Tracker on multiple RGB+X datasets. From the first to the fourth row, we jointly train Prompt Tracker on the combination of several RGB+X datasets. In the first row, we train the Prompt Tracker by only separating the embedding layers for different modalities.. In the second and third rows, we separate the CMT Prompters and the TTP Transformer layers for different modalities, respectively. In the fourth row, we separate both the CMT Prompters and TTP Transformer layers. By progressively separating the training of each module, we observe continuous improvement in performance. This phenomenon can be attributed to the limited amount of training data available for each downstream task, making it hard to train the Prompt Tracker jointly for all RGB+X tracking tasks. In the fifth row, Prompt Tracker is trained on corresponding data for specific tasks, achieving better performance.

\textbf{Temporal Matching Attention Visualization.} We visualize the temporal matching attention map of Foundation Tracker and Prompt Tracker in Figure~\ref{fig:vis}. The Foundation Tracker, after undergoing extensive large-scale training, demonstrates its strong temporal matching ability. It effectively captures the temporal dependencies and provides accurate predictions for the target object. With the parameter-efficient finetuning on downstream RGB+X tracking datasets, the Prompt Tracker further improves the tracking performance. By leveraging multimodal information and refining predictions, Prompt Tracker achieves more precise and accurate results on certain datasets. These visualizations demonstrate the effectiveness of our OneTracker in establishing temporal correspondences.

\section{Conclusion}
We propose a general framework, OneTracker, to unify several RGB tracking and RGB+X tracking tasks. OneTracker involves pretraining a Foundation Tracker on RGB tracking datasets and adapting it to downstream RGB+X tracking tasks using prompt-tuning techniques.  By leveraging the strengths of pretraining and parameter-efficient finetuning mechanisms, our framework achieves state-of-the-art results in various tracking scenarios. 
Superior performance on 11 benchmarks of 6 tasks demonstrates the effectiveness and powerful generation ability of OneTracker.

\textbf{Acknowledgments:} This work was supported by National Natural Science Foundation of China (No.62072112), Scientific and Technological Innovation Action Plan of  Shanghai Science and Technology Committee (No.22511102202).

{
    \small
    \bibliographystyle{ieeenat_fullname}
    \bibliography{main}
}

\clearpage
\setcounter{page}{1}
\maketitlesupplementary

\section{Discussion}
To the best knowledge of ours, we are the first to unify visual object tracking in a general framework. Although there exists some works~\cite{athar2023tarvis,yan2023universal,wang2023omnitracker,yan2022towards,ma2022unified,wang2021different} which tackle multiple tracking tasks in a single model, these works only consider RGB modality and ignore multimodal informtaion. Moreover, some methods~\cite{zhu2023visual,yang2022prompting} take multimodal information into consideration, but they only focus on specific modalities and still treat RGB and RGB+X tracking as separate entities. We consider these two tasks as a unified whole. Our work unifies several tracking tasks, RGB tracking and RGB+N/D/T/E/M tracking, and achieves competitive performance on 11 benchmarks across the 6 tasks.

Diverging from conventional approaches that perform full finetuning on downstream datasets, we break the widely-used full finetuning manner, and introduce the parameter-efficient transfer learning (PETL), which is popular in NLP, into tracking. In NLP, a large-scale foundation model is trained on broad data and owns a strong logical reasoning and generative ability. Then, PETL techniques are adopted to transfer foundation model to downstream tasks by freezing the pretrained weights and training inserted parameters. Due to the similar temporal matching mechanisms in RGB and RGB+X tracking tasks, we follow the large-scale training and PETL manner in NLP. Our framework begins with the pretraining of Foundation Tracker on large-scale RGB tracking datasets, enabling it to acquire a strong temporal matching ability. After that, we incorporate multimodal information as prompt and introduce CMT Prompters to enhance Prompt Tracker with multimodal features, boosting overall performance. Despite similar structure is discussed in ProTrack and ViPT, they do not take language and mask into account. Besides, TTP Transformer layers are utilized to adapt Prompt Tracker to downstream tasks better. Through adjusting a set of additional parameters (about 2.8M), Prompt Tracker inhibits the strong ability from Foundation Tracker, and have better adaptability than full finetuning. Importantly, the parameter efficiency makes it particularly suitable for resource-constrained devices where only a small number of parameters need to be distributed to end-side deployments for the generalization of downstream scenarios.

\textbf{Limitations.}
Despite the high effectiveness and efficiency, our framework still has some limitations. Firstly, for different tracking tasks within the RGB+X domain, our Prompt Tracker still needs to be trained on specific datasets separately. This implies that if we want to handle multiple RGB+X tracking tasks, we need to adjust the parameters of the CMT Prompters and TTP Transformer layers accordingly. Although the parameters of these two modules are iightweight and can be almost negligible, it still results in inconvenience. Exploring methods to handle multiple tasks within a general model through joint training is an important direction for future research. Secondly, although our model is capable of handling 6 tracking tasks across various modalities, there are still other modalities that have not been considered. We will continue to extending our model to more downstream tasks. Thirdly, as the landscape of downstream RGB+X tasks evolves, it is crucial to make our Prompt Tracker adaptive to new tasks while maintaining its original capabilities. Ensuring the flexibility of our framework to accommodate emerging tasks without sacrificing its performance on existing tasks is an important challenge that requires further investigation. Addressing these limitations will contribute to the continuous development and improvement of our framework, making it more versatile, adaptable, and effective for a broader range of tracking tasks.

\section{Experiment Details}

\begin{table*}[htbp]
	\centering
	\renewcommand\arraystretch{1.15} 
    \begin{minipage}[]{0.33\linewidth}
        \centering
        \footnotesize
        \caption{Training setting for Foundation Tracker on RGB traking datasets.}
        \renewcommand\arraystretch{1.03} 
        \begin{tabular}{c|c}
        \hline
        Config&Value\\
        \hline
        optimizer           & AdamW \cite{kingma2014adam}   \\ 
       learning rate in head           & $4\times 10^{-4}$      \\ 
       learning rate in backbone           & $4\times 10^{-5}$      \\
         weight decay             & $10^{-4}$     \\  
         batch size              &128       \\ 
                epoch              &300       \\
           learning rate decay epoch              &240       \\
          learning rate decay  factor              &10       \\
         learning rate schedule            & steplr     \\  
         maximum sampling frame gap           & 200    \\
         \hline 
\end{tabular}

        \label{tab:sup_foundation}
    \end{minipage}
    \hfill
    \begin{minipage}[]{0.33\linewidth}
        \centering
        \footnotesize
        \caption{Finetuning setting for Prompt Tracker on RGB+N/D/T/E tracking.}
        \renewcommand\arraystretch{1.13} 
        \begin{tabular}{c|c}
        \hline
        Config&Value\\
        \hline
        optimizer           & AdamW \cite{kingma2014adam}   \\ 
       learning rate       & $4\times 10^{-5}$      \\
         weight decay             & $10^{-4}$     \\  
         batch size              &128       \\ 
                epoch              &60       \\
           learning rate decay epoch              &48       \\
          learning rate decay  factor              &10       \\
         learning rate schedule            & steplr     \\  
         maximum sampling frame gap           & 200    \\
         \hline 
\end{tabular} 

        \label{tab:sup_prompt}
    \end{minipage}
    \hfill
    \begin{minipage}[]{0.33\linewidth}
        \centering
        \caption{Finetuning setting for Prompt Tracker on RGB+M tracking datasets.}
        \renewcommand\arraystretch{1.25} 
        \footnotesize
        \begin{tabular}{c|c}
    \hline
    Config&  Value\\
    \hline
    optimizer           &AdamW \cite{kingma2014adam}   \\  
  base learning rate           & $1\times 10^{-5}$      \\
     weight decay             & $1\times 10^{-7}$     \\  
     batch size              &8       \\
            Iterations              &150,000       \\ 
       learning rate decay iteration              &125,000       \\
     learning rate schedule            & steplr     \\  
     maximum sampling frame gap           & 25    \\  
   \hline
    \end{tabular}
        
        \label{tab:sup_rgbm}
    \end{minipage}
	
	\label{tab:sup_table}
\end{table*}

\subsection{Foundation Tracker Training}
Foundation Tracker are trained on a combination of several RGB tracking and object detection datasets, including LaSOT~\cite{fan2019lasot}, TrackingNet~\cite{muller2018trackingnet}, GOT-10K~\cite{huang2019got}, and COCO~\cite{coco}, following~\cite{ye2022joint,gao2023generalized,mixformer}. We only used the training sets of these dataset for training.Data augmentations, such as horizontal flip and brightness jittering, are adopted during training. 

Compared to previous trackers, the training datasets and setting, such as the number of training epochs, remain consistent. Despite the \textbf{same} training setting, our Foundation Tracker achieves superior performance, outperforming other trackers by at least 0.6 AUC on LaSOT. Models like UNINEXT and OmniTracker, which aim to address multiple vision tasks, utilze a larger set of datasets in addition to RGN tracking dataset6s. The training of UNINEXT and OmniTracker require significantly more time and GPU resources, typically taking several days and utilizing more GPUs. In contrast, our Foundation Tracker can be trained in about one day on 4 NVIDIA RTX 3090 GPUs. Compared to these models which required much more training data and training cost than our Foundation Tracker, our Foundation Tracker achieves better performance on RGB tracking (at least 1.3 AUC on LaSOT). Considering that our Foundation Tracker achieves better performance on RGB tracking while utilizing the \textbf{same or smaller} amount of training data and computational resources compared to other models, the comparison on LaSOT and TrackingNet benchmarks is both fair and favourable to our approach.

\subsection{Prompt Tracker Finetuning}
\textbf{RGB+N/D/T/E tracking.}
For the parameter-efficient finetuning of Prompt Tracker on downstream RGB+X tracking tasks, we finetune Prompt Tracker on each task separately. The size for template and search frame is $192 \times 192$ and $384 \times 384$. For RGB+N tracking, we adopt OTB99~\cite{li2017tracking}, LaSOT~\cite{fan2019lasot}, and TNL2K~\cite{wang2021towards} as training sets. For RGB+D tracking, DepthTrack~\cite{yan2021depthtrack} is chosen for training. For RGB+T tracking, LasHeR~\cite{lasher} is utilized for training. For RGB+E tracking, VisEvent~\cite{visevent} is leveraged for training. The hyper-parameters are in Table~\ref{tab:sup_prompt}. 

\textbf{RGB+M tracking.}
We choose the popular RGB+M tracking datasets, DAVIS17~\cite{pont20172017} and YouTube-VOS~\cite{xu2018youtube} for finetuning. We select the first frame and previous frame as template frame, and do not implement any cropping operation on the template and search frames. The finetuning details are in Table~\ref{tab:sup_rgbm}

\end{document}